\documentclass{article}

\usepackage{pdfx}
\usepackage{microtype}
\usepackage{graphicx}
\usepackage{subfigure}
\usepackage{booktabs} 
\usepackage{amsmath}
\usepackage{amssymb}
\usepackage{multirow}

\usepackage{hyperref}

\newcommand{\A}{\mathcal{A}}

\usepackage[accepted]{icml2023}

\usepackage{amsmath}
\usepackage{amssymb}
\usepackage{mathtools}
\usepackage{amsthm}
\usepackage{bm}
\usepackage{algorithm}
\usepackage{algpseudocode}
\usepackage{xcolor}
\usepackage{wrapfig}
\algnewcommand{\LeftComment}[1]{\Statex \(\triangleright\) #1}

\usepackage[capitalize,noabbrev]{cleveref}

\DeclareMathOperator*{\argmax}{argmax}

\newcommand{\method}{instructRL}

\newcommand{\toygame}{Say-Select}
\newcommand{\centered}[1]{\begin{tabular}{@{}l@{}} #1 \end{tabular}}

\theoremstyle{plain}

\theoremstyle{definition}

\theoremstyle{remark}

\usepackage[textsize=tiny]{todonotes}

\icmltitlerunning{Language Instructed RL for Human-AI Coordination}

\begin{document}

\twocolumn[
\icmltitle{Language Instructed Reinforcement Learning for Human-AI Coordination}




\begin{icmlauthorlist}
\icmlauthor{Hengyuan Hu}{stanford}
\icmlauthor{Dorsa Sadigh}{stanford}
\end{icmlauthorlist}

\icmlaffiliation{stanford}{Stanford University}

\icmlcorrespondingauthor{Hengyuan Hu}{hengyuan.hhu@gmail.com}

\icmlkeywords{Machine Learning, ICML}

\vskip 0.3in
]



\printAffiliationsAndNotice{}  

\begin{abstract}
One of the fundamental quests of AI is to produce agents that coordinate well with humans. 
This problem is challenging, especially in domains that lack high quality human behavioral data, because multi-agent reinforcement learning (RL) often converges to different equilibria from the ones that humans prefer.
We propose a novel framework, \textit{\method}~, that enables humans to specify what kind of strategies they expect from their AI partners through natural language instructions.
We use pretrained large language models to generate a prior policy conditioned on the human instruction and use the prior to regularize the RL objective. This leads to the RL agent converging to equilibria that are aligned with human preferences.
We show that \method~converges to human-like policies that satisfy the given instructions in a proof-of-concept environment as well as the challenging Hanabi benchmark. 
Finally, we show that knowing the language instruction significantly boosts human-AI coordination performance through human evaluations in Hanabi.

\end{abstract}

\section{Introduction}


One of the most fundamental yet challenging goals of AI is to create agents that can coordinate with humans in human-AI hybrid environments. 
In domains where abundant, high quality human behavioral data is available, such as Diplomacy~\cite{diplomacy/science}, we can expect AI agents to achieve human level performance and coordinate effectively with humans.
However, these methods are limited to settings where high quality human data is readily available. 
Without access to human data, we have to rely on techniques such as reinforcement learning (RL) algorithms to learn strong policies in multi-agent settings. 

However, the main challenge in leveraging RL for human-AI coordination is the existence of multiple, mutually incompatible equilibrium policies in a multi-agent system and the fact that without guidance RL agents can converge to any of them~\cite{ShihSKESiclr21, hu2020other}. 
Here \textit{equilibrium} policies refer to optimal or near-optimal joint policies in multi-agent environments.
In practice, humans often prefer specific subsets of policies --- particular equilibria in multi-agent games --- that align well with our capabilities and common sense, while reinforcement learning policies that do not integrate any forms of human priors often converge to policies that are hard for humans to collaborate with~\cite{bakhtin2022mastering,carroll2019utility}.

\begin{figure*}[t]
\centering
\begin{minipage}{.3\textwidth}
\centering
\includegraphics[width=1\linewidth, trim=0cm 0.cm 0cm 0cm,clip]{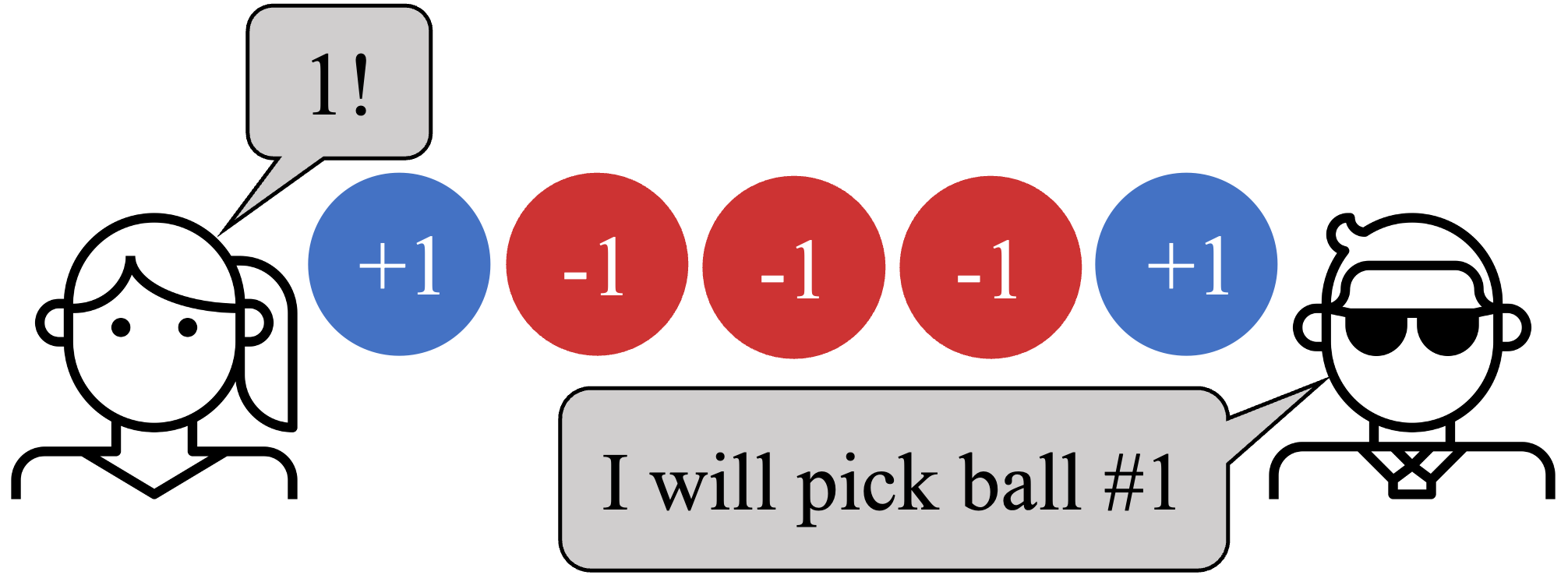}
\end{minipage}\qquad
\begin{minipage}{.3\textwidth}
\centering
\includegraphics[width=1\linewidth, trim=0cm 0.cm 0cm 0cm,clip]{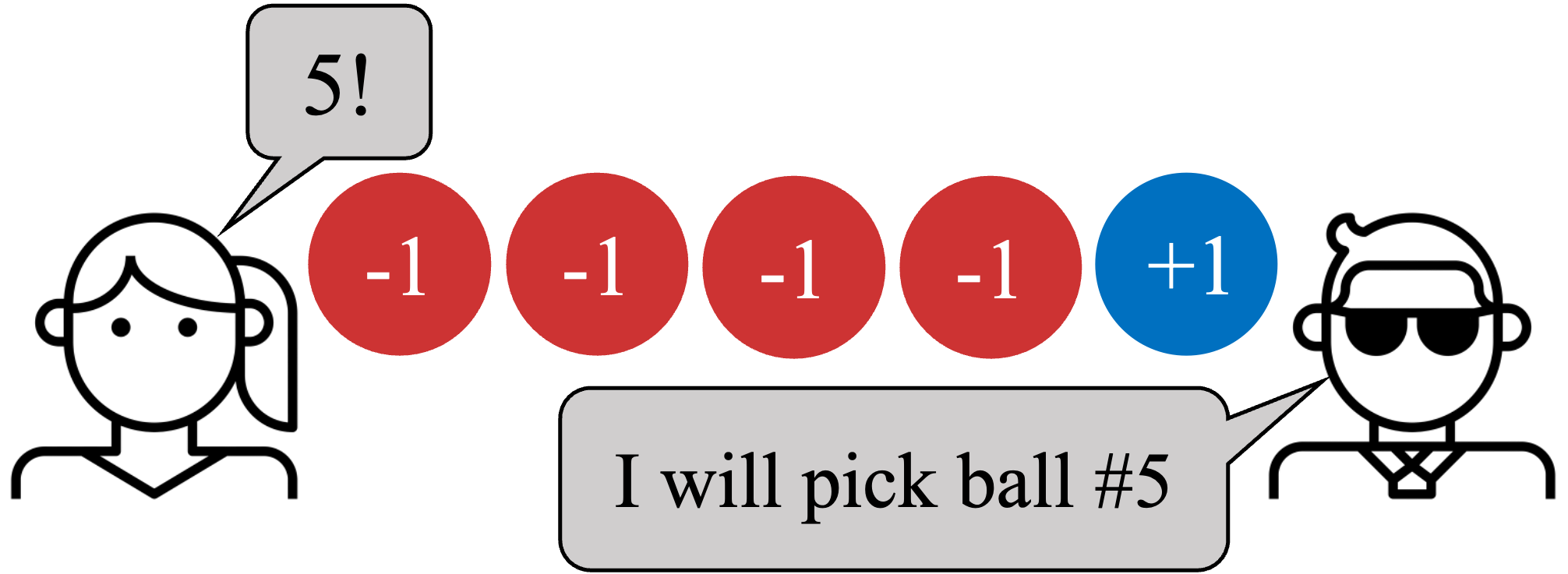}
\end{minipage}\qquad
\begin{minipage}{.3\textwidth}
\centering
\includegraphics[width=1\linewidth, trim=0cm 0.cm 0cm 0cm,clip]{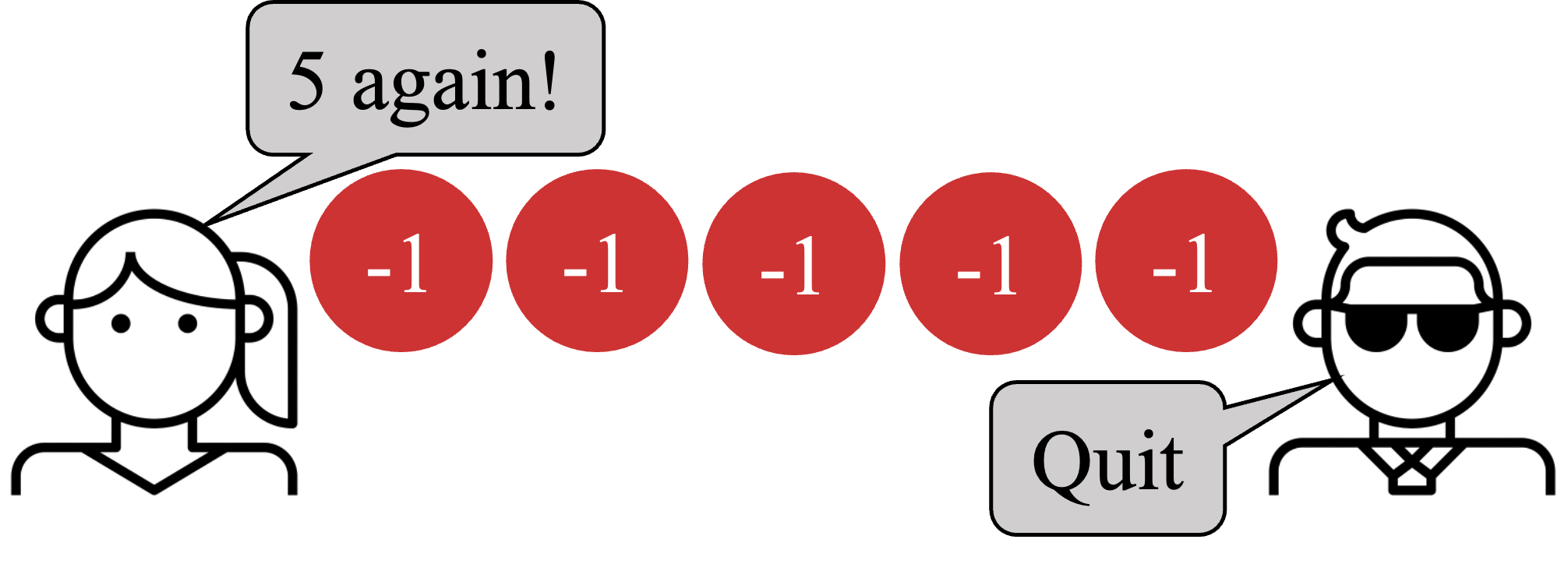}
\end{minipage}
\caption{\small Illustration of one episode of the toy example. \textbf{Left}: At the beginning of the episode, two random balls are assigned with +1 while the others are assigned with -1. Alice says `1' to Bob. Bob picks up ball \#1 and the team gets +1 reward. \textbf{Middle}: The ball is put back to the table but now assigned with -1. Alice says `5' to Bob and Bob picks up ball \#5. \textbf{Right}: Now that all the balls have -1 reward. Alice says `5' again to Bob. Bob realizes there must be no positive reward balls left, so he quits.}
\label{fig:toy-game}
\end{figure*}

Let us consider a simple collaborative game as our running example. Alice and Bob are collaborating with each other on a game ``Say-Select" as shown in Figure~\ref{fig:toy-game}.
There are 5 balls in front of Alice and Bob. A random number of them are assigned with +1 reward and the remaining ones are assigned with -1 reward. Alice (left) can see the values of the balls while Bob (right, blind-folded) cannot. They need to collaborate to collect rewards. Alice can refer to any of the balls by communicating a number from 1 to 5 to Bob. Bob can either pick up a ball or quit the game. After collecting the reward associated with the ball, the ball is re-assigned with -1 reward and put back. The game keeps running until Bob quits.
It is easy to see that from RL algorithms' perspective, there are numerous equally optimal joint policies that achieves perfect results by learning an arbitrary mapping from Alice's past action sequences to Bob's decision. However, it is unclear how existing RL algorithms can reliably produce the policies that are most natural to human. For example, a natural policy is one shown in Figure~\ref{fig:toy-game}, where when Alice communicates any number $x$, Bob picks up that ball, and if Alice communicates $x$ twice in a row, Bob should quit. 

A popular line of research attempts to address this problem by first producing a diverse set of policies and then train a common best response strategy that may generalize to any humans partner~\cite{lupu2023adversarial, pmlr-v139-lupu21a, charakorn2022generating}. These methods implicitly require the underlying RL algorithm to generate policies near the equilibria that humans prefer, which by itself is challenging and problem dependent. They also have much higher computational cost in order to produce enough policies to facilitate generalization. In \toygame, a common best response trained against all possible optimal policies still needs to spend many episodes exploring and identifying its partner's policy when paired with a human. Meanwhile, humans may also adjust their policy in parallel as they try to understand and adapt to the best response, which makes the problem more complicated.

Our work seeks to address this challenge based on two key observations. First, humans can better understand and coordinate with a policy if it can concisely be summarized in natural language. Second, in most real world coordination scenarios, humans talk to each other or even negotiate to achieve some agreements on how they should collaborate, i.e., what conventions or equilibrium to follow. 
For example, in \toygame, a human can just tell the RL algorithm to produce joint policies \emph{where Bob selects the same number as Alice does}, eliminating coordination overhead when deployed to play with the human.
These two important aspects have not been considered in the existing multi-agent reinforcement learning (MARL) methods and we aim to incorporate them for the goal of guiding RL agents towards more human-like policies.

In this paper, we propose a novel framework for human-AI coordination where the human can provide high-level natural language instructions to the AI partner as additional specification so that the agent learns to coordinate with the human in ways that match the human's expectation. 
This would guide the AI agent to follow human's instructions during RL training and agree on a specific equilibrium to converge to.
The key idea of our approach \textit{instruct-RL} is to leverage large language models (LLMs)~\cite{gpt3} to regularize multi-agent RL training process based on the provided human instructions.
We first construct a prior policy by querying large language models (LLMs) given the instructions and short descriptions of the current observation.
Then we use the LLM prior to regularize an RL algorithm such as Q-learning~\cite{dqn} or PPO~\cite{ppo} so that the converged equilibrium satisfies the instructions.

We initially evaluate our method in the purposely designed \toygame~game discussed above, where we show that our method learns intuitive, human-compatible policies as instructed by the language instructions. 
Then, we evaluate our method in the large scale Hanabi benchmark~\cite{bard2020hanabi}. We show that we can obtain equally strong but qualitatively different policies given different instructions and humans can better coordinate with the AI agents when they know the language instructions that the agents follow.


\section{Related Work}

\textbf{Language and Human Inductive Bias:} Our research is closely related to the works that study the role of natural language in learning. On the relationship between natural language and the inductive bias of human learning, ~\citet{kumar2022using} show that training RL agents with the auxiliary tasks of predicting representations from human-generated natural language task descriptions leads to more human-like behavior. Their conclusion aligns well with our motivation that human's inductive bias makes us prefer policies that can be easily described in language.

\textbf{Language for Exploration in RL:} Natural language has also been used to improve RL in various ways. The most relevant ones use language abstraction for exploration. \citet{ella} rewards the agent for finishing any semantically meaningful low-level such as `picking up a key' and gradually build up a dataset for learning to correlate the high-level instructions with low-level descriptions.
\citet{semantic-exploration-lang} and \citet{mu2022improving} use language or visual language models to discover novel states for intrinsic rewards at the semantic level. These works focus on learning a better policy in single agent RL setting by addressing the hard exploration problem. In comparison, our work focuses on the \textit{equilibrium selection} problem --- converging to a human-like equilibrium --- in multi-agent systems. These two problems are orthogonal. Even in environments where exploration is not an issue, such as our Say-Select example, we still need novel methods to obtain a human-like policy.

\textbf{Foundation Models and In-Context Learning for RL:} 
Prior works have explored using foundation models for reward specifications in RL~\cite{llm-reward, minedojo}.
The in-context learning capability of the large foundation models~\cite{gpt3, fm} allows user to specify reward function with natural language descriptions. 
MineDojo~\citep{minedojo} collects a large scale multi-modal Minecraft dataset from the internet and trains a CLIP~\cite{clip} style contrastive video-text model. 
It then uses the video-text model to provide dense reward for RL by computing the similarity between the embedding of in-game video clip and the embedding of the language description of the task. 
Apart from our focusing on the \textit{equilibrium selection} problem  mentioned above, our method is also different in that it does not rely on purposely collected domain specific data but use an off-the-shelf large language models (LLMs) like GPT-3 together with the reward from the environment.

\citet{llm-reward} utilize the in-context learning capability of LLM to design hard-to-specify reward functions for RL, such as versatility, fairness etc.
Because the rewards from LLM is the only learning signal, their method can only be applied to small settings with few timesteps because LLMs need to understand the entire game logic and past history to make decision. 
In contrast, our work focuses on the settings where environment reward is crucial for learning optimal policies and LLMs help to steer the learning process to produce a policy that satisfy the language instruction. Thanks to this hybrid setup, we can apply it to more complex environments like Hanabi.

\textbf{Human-AI Coordination:} 
There are three related research directions in human-AI coordination problems. 
The first one is to use human data to directly model human behavior~\cite{carroll2019utility} or to regularize RL/search so that it learns expert level policies while staying close to human equilibria~\cite{lerer2018learning, diplomacy/science}. 
The second direction is to design cognitively inspired learning algorithms~\cite{laidlaw2022the, hu2021off, cui2021klevel} to produce more human-like policy than vanilla RL does. 
The final one seeks to produce a diverse set of policies~\cite{pmlr-v139-lupu21a, lupu2023adversarial, charakorn2022generating, strouse2022collaborating} and train a common best response to them so that it may generalize better to unseen partners.

Our work does not require human dataset but instead uses human specified language instructions and LLMs to train RL policies that matches human's preference for better coordination. It can be combined with methods from the second category to train more human-like policies that satisfies the instruction. As we show later in the experiment section, our method can also produce semantically different policies given different instructions, making it a compelling method for the third paradigm.

\section{Background}

\textbf{MARL:} 
Multi-agent RL (MARL) is a powerful tool for learning strong agents in multi-agent environments.
The environment consists of the state space $\mathcal{S}$, $N$ agents with their respective action space $\bm{\mathcal{A}} = \A^{1} \times \dots \times \A^{N}$. 
A transition function $\mathcal{T}: \mathcal{S} \times \bm{\mathcal{A}} \rightarrow \mathcal{S}$ defines the dynamics of the environment. 
We consider the fully cooperative setting where a reward function $r: \mathcal{S} \times \bm{\mathcal{A}} \rightarrow \mathbb{R}$ returns a \textit{common payoff} shared by all players given state $s$ and joint action $\bm{a}$.
We focus on the partially observable setting where each agent has their own observation function $\Omega: \mathcal{S} \rightarrow \mathcal{O}$.
Although the coordination challenge exists in fully observable settings, it is more prominent under partial observability because the inference of other agents' true intention is much harder.
The policy for each player $\pi^{i}(a^{i}_t | \tau_t^{i})$ takes in an \textit{action-observation history} $\tau_t^{i} = \{\Omega^i(s_1), a_1, ..., a_{t-1}, \Omega^i(s_t)\}$ and outputs a distribution over its action space $\mathcal{A}^{i}$. A joint policy is simply the collection of policies for all players $\bm{\pi} = (\pi^1, \dots, \pi^N)$.
The goal of MARL is to train a joint policy that achieves the maximum total return $J(\bm{\pi}) = \mathbb{E}_{\tau\sim\bm{\pi}} R(\tau)$ where $R(\tau) = \sum_{t=t_0}^{T} \gamma^t \cdot r_{t}$ is the total discounted return of a game.
$\gamma$ is the discounting factor.




\textbf{LLMs:} Large language models (LLMs)~\cite{gpt3} are generative text models trained on enormous datasets collected from Internet to predict next token given the context. 
With proper prompts~\cite{cot, promot-step-by-step}, LLMs have demonstrated impressive zero-shot or few-shot generalization capabilities on challenging tasks such as reasoning and arithmetic. 
Researchers have further fine-tuned LLMs using RL with human feedbacks (RLHF)~\cite{instruct-gpt} or instructions~\cite{flan} to generate more consistent, higher quality results.

\section{Method}

In this section, we introduce \textit{instructRL}, a language augmented RL method that can converge to different equilibria given a language instruction.
Specifically, we are interested in operationalizing instructRL in multi-agent settings for collaborating with humans. 
In such settings, the human partner can provide a language instruction guiding the joint equilibrium of the human-AI team.

The core idea of instructRL is to first construct a prior policy using LLMs that is conditioned on two pieces of information: 1) the initial human instruction and 2) a language prompt that describes relevant observations based on rolling out the current policy in the environment.
For instance, for our Say-Select example in Figure~\ref{fig:toy-game}, we can give instruction \emph{``Select the same number as Alice''} to the Bob agent and the description of the current observations can be \emph{``Alice said 1.''} 
We then train an RL agent, where its objective is regularized with the generated LLM prior as a reference policy. 

\textbf{LLM Prior Policy:}
We construct the prior policy by letting an LLM to predict the probability of possible actions given the observation and the instruction. 
To do so, we essentially need to evaluate $p_{\texttt{LLM}} [\texttt{lang}(a_t) | \texttt{lang}(\tau^i_t), \texttt{inst}]$ for all possible action at each time step, which requires language descriptions of observations $\texttt{lang}(\tau^i_t)$ and actions $\texttt{lang}(a)$.
We note that the language instruction \texttt{inst} \textit{stays fixed} during training, i.e., no active human feedbacks during the actual RL loop.

The LLM observes the game through the language descriptions of the observations.
In board game environments such as Hanabi and Diplomacy, the language description of current observation $\tau^i_t$ can be generated automatically with simple rule-based programs. 
In fact, the language observations are often the most natural medium on which humans reason about these games. 
However, in real world scenarios that require grounding in the physical environment, we may use image captioning models~\cite{clip} to generate descriptive languages of the scenes or extend our framework to allow humans to specify instructions in video format. We leave this for future explorations.

In order for the LLM to evaluate whether an action is plausible given the instruction \texttt{inst} and the current observation $\texttt{lang}(\tau^i_t)$, we must also map each action to a language description ($\texttt{lang}(a)$). It is easy to convert actions to language descriptions in the games we consider in this paper. Prior works have also assigned language descriptions to high level robotics primitives so that they can use LLMs for task level planning in real world~\cite{saycan2022arxiv, huang2022llm-planner}. However, it is worth noting that many environments contain actions that cannot be easily abstracted in language, e.g., in a robotics setting, where the actions are the continuous joint angles of a robot arm. We note that this is a limitation of our work, but we are hopeful that with the active development of new multi-modal foundation models, the ideas of this work can be applied more broadly beyond LLMs --- enabling humans to effectively guide AI agents to reach more desirable equilibria.

Having defined \texttt{inst}, $\texttt{lang}(\tau^i_t)$ and  $\texttt{lang}(a_t)$, we can compute $p_\texttt{LLM} = \textsc{Softmax}(\beta \cdot \text{logit})$ where $\beta$ is an optional scaling factor and the logit is a function of the language components, i.e. $\text{logit} = f(\texttt{inst}, \texttt{lang}(\tau^i_t), \texttt{lang}(a_t))$. For actions that have homogeneous descriptions, such as in \toygame, the logit function $f$ can simply be the prediction loss of the $\texttt{lang}(a_t)$ conditioned on some prompt that combines \texttt{inst} and $\texttt{lang}(\tau^i_t)$. Another option, which is later used in the Hanabi experiments, is to construct a question-answering style prompt that ask whether we should do $\texttt{lang}(a_t)$ given the instruction \texttt{inst} and current observation $\texttt{lang}(\tau^i_t)$. We set the logit to 1 if the probability of generating affirmative answers is greater than the probability of generating negative answers, and 0 otherwise.

The LLM prior $p_\texttt{LLM}$ itself is not sufficient to solve complex tasks.
For example, a moderate-sized LM with roughly 6B parameters can not figure out when to quit in ~\toygame in Figure~\ref{fig:toy-game} and even the largest LM to date cannot play Hanabi in human level. 
Therefore, we still need to rely on RL to find good policies, but we guide the RL policy using the LLM prior described in this section as a regularizer to satisfy the language instructions provided by the human. 

\begin{figure}
    \centering
    \includegraphics[width=0.99\linewidth, trim=0.7cm 2.3cm 5.3cm 0.85cm,clip]{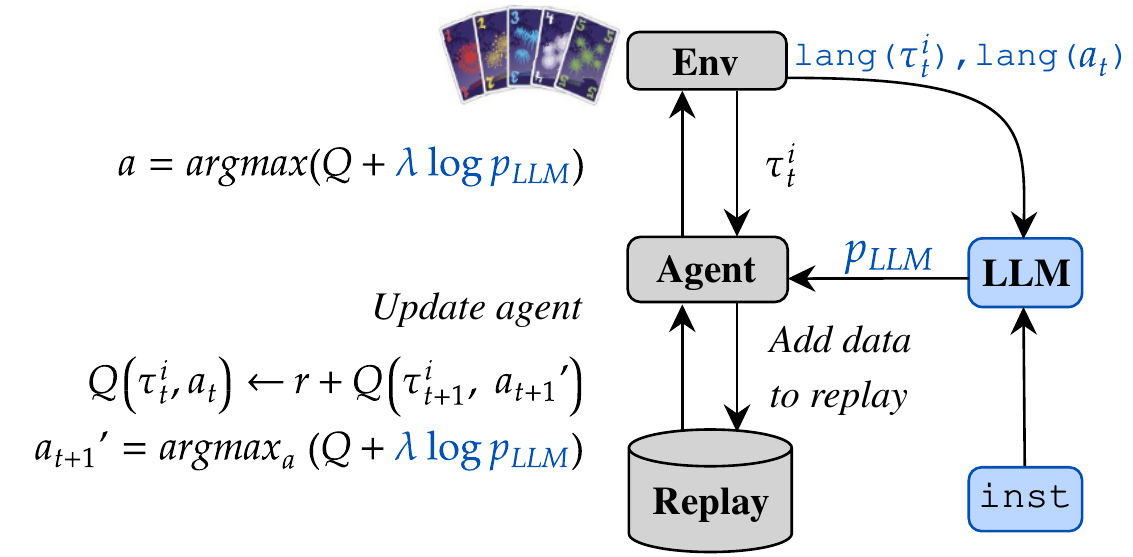}
    \caption{\small InstructQ. The differences between instructQ and normal Q-learning is highlighted in blue.}
    \label{fig:instruct-q}
    \vspace{-4mm}
\end{figure}

\textbf{Regularized RL:}
Regularization has been widely used in RL to encourage exploration~\cite{a3c}, or to encourage an RL policy to stay close to a given prior~\cite{ziebart2008maximum}. 
Notably, regularizing RL policies towards a behavioral cloned policy trained on a massive human dataset was critical for mastering complex games such as Starcraft~\cite{vinyals2019alphastar} and Diplomacy~\cite{bakhtin2022mastering}. 
In this paper, we would like to similarly regularize an RL agent to guide the equilibria towards desirable behaviors. 
Instead of relying on massive amount of human data or hand-designed reward shaping, we enable the user to regularize the RL policy by providing language instructions. The LLM prior $p_{\text{LLM}}$ effectively captures the human preferences, which allows us to instruct an RL agent towards human preferences. 
We thus refer to our algorithm as instructRL. 
We consider two types of regularization techniques for Q-learning and PPO respectively. 

For Q-learning~\cite{dqn}, we can simply augment the policy with $\log p_{\texttt{LLM}}$. 
The exploration policy becomes $a = \epsilon{-}\textsc{Greedy}(Q_\theta + {\lambda \log p_{\texttt{LLM}}})$ and the training becomes
\begin{gather*}
Q_\theta(\tau_{t}^i, a_t) \gets r_t + \gamma Q_{\theta}(\tau_{t+1}^i, a_{t+1}'), \\
\text{where } a_{t+1}' = \argmax_{a}  [Q(\tau_{t+1}^i, a) + \lambda \log p_\texttt{LLM}(\tau_{t+1}^i, a)].
\end{gather*}
We refer to this version of instructRL as \textit{instructQ}. 

For PPO~\cite{ppo, reg-ppo}, we add a KL penalty term to the objective 
$J(\theta) = \mathbb{E}_{\tau\sim\bm{\pi}_{\theta}} R(\tau) + \lambda \text{KL} (\pi_\theta || p_{\textsc{LLM}})$. The policy loss becomes
\begin{gather*}
\mathbb{E}_{\tau\sim\bm{\pi}_{\theta}} \sum_t [-\log \pi_\theta(a_t|\tau^i_t) A_t + \lambda \text{KL} [\pi_\theta(\tau^i_t) || p_{\texttt{LLM}}(\tau^i_t)]]
\end{gather*}
where $A_t = Q(\tau^i_t, a_t) - V(\tau^i_t) $ is the advantage. The value loss remains unchanged.
We refer to this as \textit{instructPPO}.


\begin{algorithm}
\caption{InstructRL (for player $i$). \textsc{Reg-Act} and \textsc{Reg-Train} functions depend on the specific regularized RL algorithm. Practical implementation uses parallel training and data collection workers.}
\label{alg:instruct-rl}
\begin{algorithmic}
\LeftComment{\texttt{inst}: language instruction specified by by human};
\LeftComment{$\texttt{LLM}$: pretrained large language model};
\LeftComment{\textsc{Construct-Prior}: function to compute LLM prior};
\LeftComment{\textsc{Reg-Act} \& \textsc{Reg-Train}: functions depends on the specific regularized RL algorithm;}
\Procedure {InstructRL} {\texttt{inst}}
\State initialize $\pi_\theta$
\While {not converged$(\pi_\theta$)}
    \State $\tau^i  \gets \{\}$ 
    \For {$t \gets 1, \dots, T$}
        \State $\tau^{i} \gets \textsc{Append}(\tau_i, \Omega^{i}(s_t))$
        \State $p_\texttt{LLM} \gets \textsc{Construct-Prior}(\texttt{LLM}, \texttt{inst}, \tau^i)$
        \State $a^{i}_t \gets \textsc{Reg-Act}(\pi_{\theta}(\tau^i), p_\texttt{LLM})$
        \State $r_t, s_{t+1} \gets \mathcal{T}(s_t, \bm{a}_t) $
        \State $\tau^{i} \gets \textsc{Append}(\tau_i, \bm{a}_t, r_t)$
    \EndFor
\State update policy $\pi_\theta \gets $ \textsc{Reg-Train}$(\pi_\theta, \tau^i)$
\EndWhile
\State \textbf{return} $\pi_\theta$
\EndProcedure
\end{algorithmic}
\end{algorithm}

We summarize our method in Algorithm~\ref{alg:instruct-rl} and provide an illustration of the instructQ version in Figure~\ref{fig:instruct-q}. In addition, we note that other regularization techniques, such as soft Q-learning~\cite{soft-q} or modifying the reward function to be $r - \lambda \log\frac{\pi_\theta(a_t|\tau^i_t)}{p_{\texttt{LLM}}(a_t|\tau^i_t)}$~\cite{rlhf}, also fit into our instructRL framework.

\section{Experiment}
In this section, we will demonstrate instructRL in two multi-agent coordination game settings: \toygame~in Sec.~\ref{sec:toy} and Hanabi in Sec.~\ref{sec:hanabi}. We will open-source code and models for both experiments.

\subsection{\toygame~ Experiment}
\label{sec:toy}
\begin{figure}[t]
    \centering
    \includegraphics[width=\linewidth]{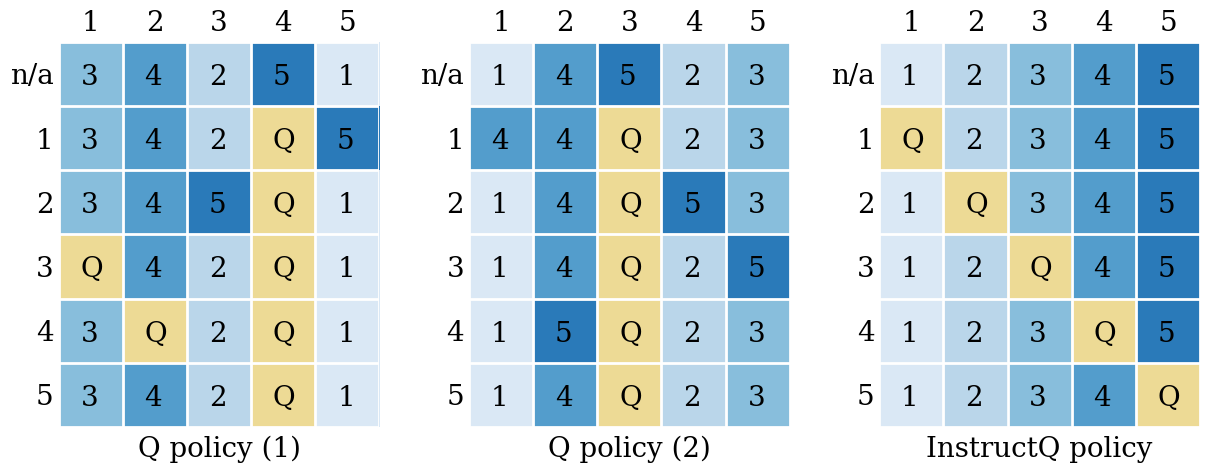}
    \vspace{-7mm}
    \caption{\small Bob's policy trained with different methods. Row values are Alice's actions \emph{two steps ago} and column values are Alice's actions \emph{one step ago}. The value in each cell is Bob's action when observing Alice's past two actions.
    Here Bob's actions are 1 through 5 (shown in different shades of blue) for selecting different balls and ``Q" (shown in yellow) refers to Bob quitting. \textbf{Left} and \textbf{Middle}: Two policies from vanilla Q-learning but with different seeds. \textbf{Right}: Policy from instructQ with $\lambda=0.25$. We note that all three policies shown here are optimal in self-play, but only the InstructQ policy is the intuitive policy that follows \texttt{inst}= ``I should select the same number as my partner''.}
    \label{fig:toy-result}
\end{figure}

We first evaluate our method on \toygame~game shown in Figure~\ref{fig:toy-game}.
In this game, as discussed earlier, an intuitive solution for Bob is to pick up the ball $\#k$ if Alice says $k$ and for Bob to quit if Alice says a number that has appeared before. This is because if Alice repeats a number, the corresponding ball will have a negative reward, and thus any reasonable partner would not repeat a number twice if there are other playable balls with positive reward. 
The observation of Alice is a tuple containing the reward of every ball and the previous action of Bob. 
Alice's action space is saying a number from 1 to 5 referring to any of the 5 balls.
It is sufficient to let Bob's observation to be the last two actions of Alice, as that would be enough for Bob to optimally respond to Alice, e.g., converge to the policy described earlier.

In this game, we use instructQ for Bob, while allowing Alice to use vanilla Q-learning. 
We set the instruction for Bob \texttt{inst} = ``I should select the same number as my partner".
We map Bob's observation to text $\texttt{lang}(\tau^i_t)$ by converting Alice's \textit{most recent} action (1 through 5) from integer to string. 
Note that the RL policy still observes the last two actions. 
For Bob's actions, we map them to string $\{``0", ``1", \dots ,``5"\}$ with 0 for quitting and the remaining 1 through 5 for selecting the corresponding ball.
We combine all these components to create the following prompt: 
\begin{quote}
``\{\texttt{inst}\}.\\
\emph{My partner selected} $\texttt{lang}(\tau^i_t)$. \\
\emph{So I should select}"
\end{quote}
We feed the prompt to an open-sourced GPT-J~\cite{gpt-j} model with 6 billion parameters and use the prediction loss for the action strings as logits~\cite{saycan2022arxiv}. Finally, we apply \textsc{Softmax} to the logits with $\beta=1$ to get the prior policy $p_\texttt{LLM}$.
We use tabular Q-learning with no neural network as the state space is small enough and regularization weight $\lambda=0.25$ for instructQ. Details on the hyper-parameters are in Appendix~\ref{app:toy}.
\begin{table*}
    \centering
    \begin{tabular}{l l}
    \toprule
      Action type   & Text observation example \\
    \midrule
      Null          &  My partner did nothing \\
      Discard       &  My partner discarded their card at position `A' \\
      Play          &  My partner played their card at position `B' \\
      Hint rank     &  My partner told me that the rank of my card at position `D' is two \\
      Hint color    &  My partner told me that the color of my cards at positions `A' and `C' is red \\
    \bottomrule  
    \end{tabular}
    \caption{\small Examples of text observation $\texttt{lang}(\tau^i_t)$ for partner's all 4 types of past actions. We use position A,B,C,E,D to refer to different cards in hand with A being the oldest position. Null means that it is the beginning of the game.}
    \label{tab:hanabi-text-obs}
\end{table*}

\begin{table}
    \centering
    \begin{tabular}{l l}
    \toprule
      Action type   & Language description example \\
    \midrule
      Discard       &  discard my card at position `B' \\
      Play          &  play my card at position `A' \\
      Hint rank     &  hint rank to my partner \\
      Hint color    &  hint color to my partner \\
    \bottomrule  
    \end{tabular}
    \caption{\small Examples of language descriptions for actions $\texttt{lang}(a_t)$. Note that we map all 5 hinting rank actions to the same language (same for hinting color actions).}
    \vspace{-5mm}
    \label{tab:hanabi-text-action}
\end{table}

\textbf{Result:} 
With instructQ, Alice and Bob always (10 out of 10 seeds) converge to the intuitive joint policy. 
In the converged joint policy, Alice says a number that corresponds to a $+1$ reward ball if there any balls with positive reward are still left and repeats her most recent action otherwise, i.e., if there no $+1$ reward balls are left. 
Bob quits if the last two actions from Alice are the same. 
Otherwise, he selects the ball with the same number as Alice says. Bob's policy is illustrated in the right plot of Figure~\ref{fig:toy-result}. 
On the left and middle shows two policies from normal Q-learning, i.e., same hyper-parameters except for $\lambda=0$. Although all three policies are \textit{optimal} in self-play, the instructQ policy is clearly easier for human to coordinate with. The language instruction associated with the instructQ policy makes it effortless to understand Alice's intents even without any explicit communication.

\subsection{Hanabi Experiment}
\label{sec:hanabi}

Hanabi is a fully cooperative card game that is often used as a benchmarking environment for MARL and human-AI coordination~\cite{bard2020hanabi}.
It requires implicit communication through actions and reasoning about other people's intentions.
In this section, we first introduce the rules of Hanabi, followed by two different strategies that are commonly used by humans.
Then we show that both instructQ and instructPPO can produce these two strategies given their corresponding language instructions.
Finally, through human evaluation, we show that humans and AI can coordinate much more effectively when they know the instructions used in training of the RL agents.
They also think that the agents' policies satisfy the language instructions and the language helps them better coordinate with the agents.

\textbf{Hanabi rules and strategies:}
We consider the 2-player version of Hanabi in this paper.
The deck in Hanabi consists of 5 color suits. Each suit has 10 cards divided into 5 ranks with three 1s, two 2s, two 3s, two 4s and one 5. 
The players need to collaborate to play exactly 5 cards of each color in increasing order of rank. 
For example, when no cards have been played, the 1s of all colors are valid play. If a red 1 has been played, then the red 2 or the 1s of other colors are valid play, etc.
Each player maintains five cards in their hands. They can see their partner's cards but not their own, so they need to use hints to inform their partner.
Players take turns to move. In each turn, the active player can either play a card, discard a card, or hint a color/rank to their partner, which informs the recipient which cards in their hand have that specific color/rank. The game terminates if the player plays an unplayable card for 3 times or when the deck is exhausted. The final score of the team is 0 in the first case and equals to the number of cards played in the second case. The maximum score is therefore 25.

The strategies in Hanabi are centered around giving hints.
The game starts with 8 hint tokens.
Players consume a hint token when they hint and regain a token when they discard.
Therefore, they often need to discard cards at certain pace to get playable cards and also recoup hint tokens.
Because there are only limited copies of each card value, players need to use hint not only to tell their partner which card to play, but also which card to save.
For example, if our partner is holding the only red 3 left in the game but red 3 is currently not playable, we may want to tell them the importance of this card by either hinting color red or hinting rank 3 to them.
The dual purpose of the hint actions make the game challenging because they may be misinterpreted, especially when the card we want to save shares some property with other cards in the partner's hand. A common strategy that experienced players use is to designate hinting colors to indicate playable cards and hinting ranks to indicate cards that need to be saved. We refer to this as the \textit{color-based} policy. Alternatively, we can swap the role of hinting color and hinting rank to get a different but equally reasonable \textit{rank-based} policy. See Appendix~\ref{app:hanabi-illu} for a more intuitive illustration with a picture of an actual game.

\textbf{Setup:}
First, we show that \method~can produce the two policies mentioned above with different language instructions. 
Specifically, we use \method~with instruction \texttt{inst\_color} = \textit{``If my partner tells me the `color' of some of my cards, I should `play' those specific cards. If my partner does something else, e.g. discards their card or tells me the `rank' of my cards, then I may `hint color' to my partner"} to produce the color-based policy.
By swapping \textit{``rank"} for \textit{``color"} and \textit{``color"} for \textit{``rank"}, we get \texttt{inst\_rank} and use it produce the rank-based policy. 
We use the 175B parameter GPT-3.5 named text-davinci-003 as our LLM.

The observation in Hanabi contains lots of information such as the partner's hand, which cards have been played or discarded, previous actions, etc.
It is challenging for LLMs to digest language descriptions of the entire observation.
Since our instructions mainly capture the high level relationships between partner's actions and our response, we design the language description function $\texttt{lang}(\tau^i_t)$ to return the partner's last action in text.
Table~\ref{tab:hanabi-text-obs} shows examples of text observations $\texttt{lang}(\tau^i_t)$ for all types of past actions that our agent may observe.
We use letters A, B, C, D and E instead of 1st, 2nd, 3rd, 4th and 5th to refer to different hand positions to avoid spurious correlations between card positions and ranks.
Similarly, it is straightforward to convert the actions to language descriptions.
Examples of $\texttt{lang}(a_t)$ are shown in Table~\ref{tab:hanabi-text-action}.


We combine the instruction, observation and action together using the following prompt:
\begin{quote}
\emph{Instruction:} \{\texttt{inst}\}. \\
\emph{Previously:} \{\texttt{lang}$(\tau^i_t)$\}. \\
\emph{Question: Should I }\{\texttt{lang}$(a_t)$\}? \\
\emph{Answer:} 
\end{quote}
Then we ask GPT-3.5 to predict the next token and set the logit of the action to be 1 if $p(\text{yes}) > p(\text{no})$ and 0 otherwise.
The question answering style template is inspired by NLP works~\cite{flan, cot} that study how to prompt LLMs effectively. 
We pre-compute the logits for all \texttt{lang}$(\tau^i_t)$ and \texttt{lang}$(a_t)$ pairs and cache them before training RL. 
During the RL training, we compute the prior policy $p_{\texttt{LLM}} = \textsc{Softmax}(\beta \cdot \text{logit})$ over the legal actions at each time step. 
The total GPT API bill for this project, including the cost for tuning the instructions and cost for additional studies in the Appendix, is roughly US\$200.

Note that we set the logit to 0 or 1 instead of using the actual probability of generating the action text for two reasons. 
First, we do not need a fine-grained prior policy from the LLM because it does not take the game rule as input and only observes limited information. 
The LLM prior policy only needs to provide a coarse guidance to bias the RL agent to converge to desired equilibria.
Second, we do not want the RL agent to suffer from the inherent biases of the LLM because they can lead to sub-optimal outcomes. 
For example, under \texttt{inst\_rank}, if previously my partner told me that the rank of my card A and C is 1, then LLM gives higher probability for yes if the action is to play card B ($4\%$) than if the action is to play card E, F ($<0.01\%$).

In Appendix~\ref{app:robust}, we include a more detailed analysis on the instructions by comparing the LLM predictions with a rule-based oracle in~\ref{app:robust-acc}.
We also have robust analysis for instructRL with simpler instructions and noisy LLM priors in Appendix~\ref{app:robust-simple} and~\ref{app:robust-noise} respectively.

We implement instructQ and instructPPO based on the open sourced repository of off-belief learning (OBL)~\cite{hu2021off}. The most noteworthy implementation detail is that we train all our models by fine-tuning an OBL level 1 policy instead of training from scratch and we anneal $\lambda$ as training progresses. The Q-learning and PPO baselines use the same setting except for $\lambda=0$.
More discussions on this and other implementation details is in Appendix~\ref{app:impl-hanabi}.

\begin{table}[t]
    \centering
    \begin{tabular}{l c c }
    \toprule
      Method              & Self-play & Intra-AXP \\
      \midrule
      Q-learning          & 23.96 $\pm$ 0.05 & 23.77 $\pm$ 0.07  \\
      InstructQ (color)  & 23.78 $\pm$ 0.05 & 23.77 $\pm$ 0.06 \\
      InstructQ (rank)   & 23.92 $\pm$ 0.02 & 23.78 $\pm$ 0.05 \\
      \midrule
      PPO                  &  24.25 $\pm$ 0.01  & 24.25 $\pm$ 0.01 \\
      InstructPPO (color) &  24.25 $\pm$ 0.03  & 24.23 $\pm$ 0.01 \\
      InstructPPO (rank)  &  24.10 $\pm$ 0.02  & 24.08 $\pm$ 0.02 \\
     \bottomrule
    \end{tabular}
    \caption{\small Performance of different algorithms. We run 3 different seeds for each method. \textbf{Self-play}: evaluate the agents with themselves. \textbf{Intra-AXP} (intra-algorithm cross-play): evaluate pairs of agents trained with the same method but different seeds. A small gap between Intra-AXP and self-play indicates that the learning algorithm consistently converges to roughly the same solution.}
    \label{tab:hanabi-xp}
\end{table}

\begin{figure}[t]
    \centering
    \includegraphics[width=\linewidth]{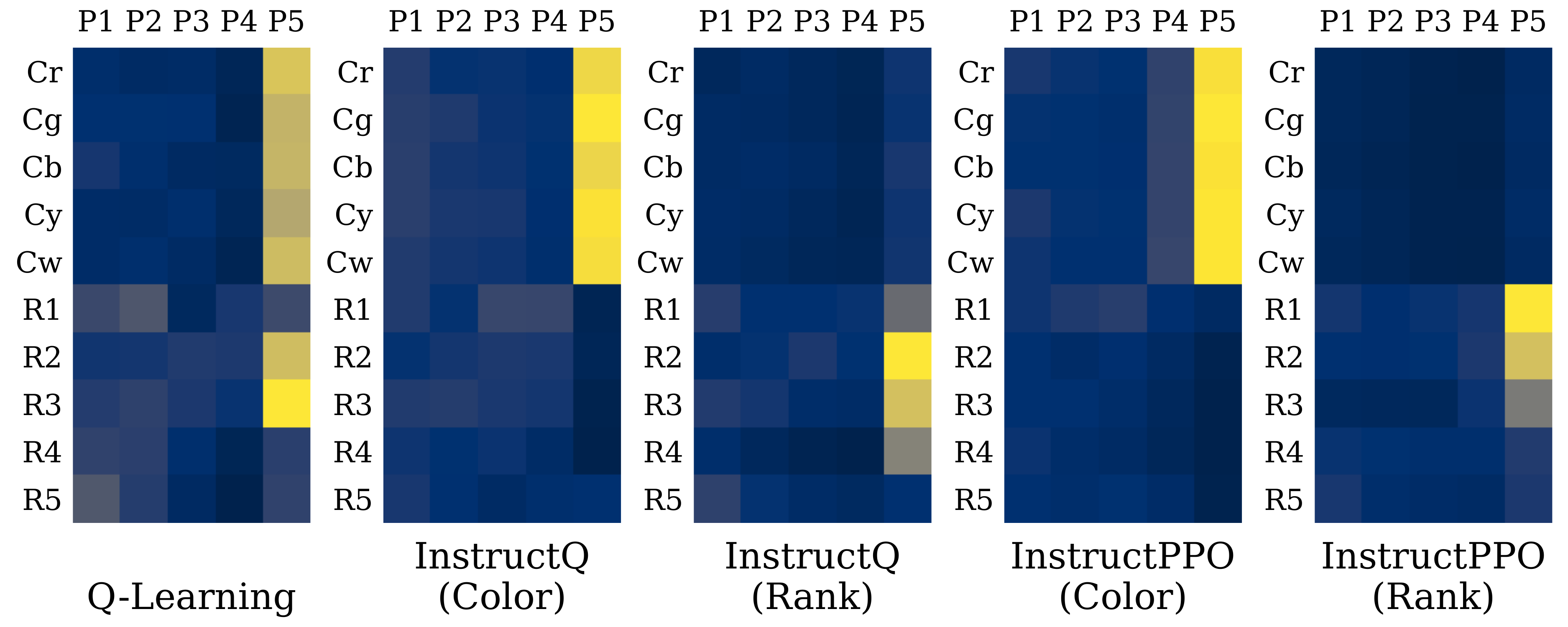}
    \vspace{-5mm}
    \caption{\small Conditional action matrix  $p(a_{t+1} | a_t)$ for different agents. We only show most relevant action pairs for conciseness. The row values are the actions from the active player at time step $t$. \textbf{Cr} through \textbf{Cw} correspond to the action of hinting color red, green blue yellow and white respectively. \textbf{R1} through \textbf{R5} correspond to the actions of hinting rank 1 through 5. The column values are the actions from the active player at time step $t+1$. \textbf{P1} through \textbf{P5} correspond to playing the card at position 1 through 5 with 5 being the newest position. 
    For each cell $p_{(i,j)}$, we first count all occurrences of the action pair over 1000 games and then normalize it $\sum_{i,j} p_{i,j} = 1$. 
    Bright yellow means high probability and dark blue means low probability. 
    All the policies focus on playing their newest cards but they demonstrate different hinting strategies.
    }
    \label{fig:action_matrix}
\end{figure}

\textbf{Results:}
We first investigate the performance and reproducibility of instructQ and instructPPO with both color and rank instructions.
From Table~\ref{tab:hanabi-xp}, we see that under both color and rank instructions, two instructRL variants achieve similar self-play (evaluating agents with themselves) and intra-AXP (intra-algorithm cross-play: evaluating pairs of agents trained with the same method but different seeds) scores as their vanilla counterparts. Intra-AXP is an important sanity check because a method is unlikely to work well for human-AI coordination if the agents trained by the method cannot coordinate with other agents trained with the same method but different seeds.
The small gap between Intra-AXP and Self-play of the tested method indicates that they converge to nearly identical policies across different seeds, meeting the precondition for human-AI coordination.

\begin{figure}
    \centering
    \includegraphics[width=\linewidth]{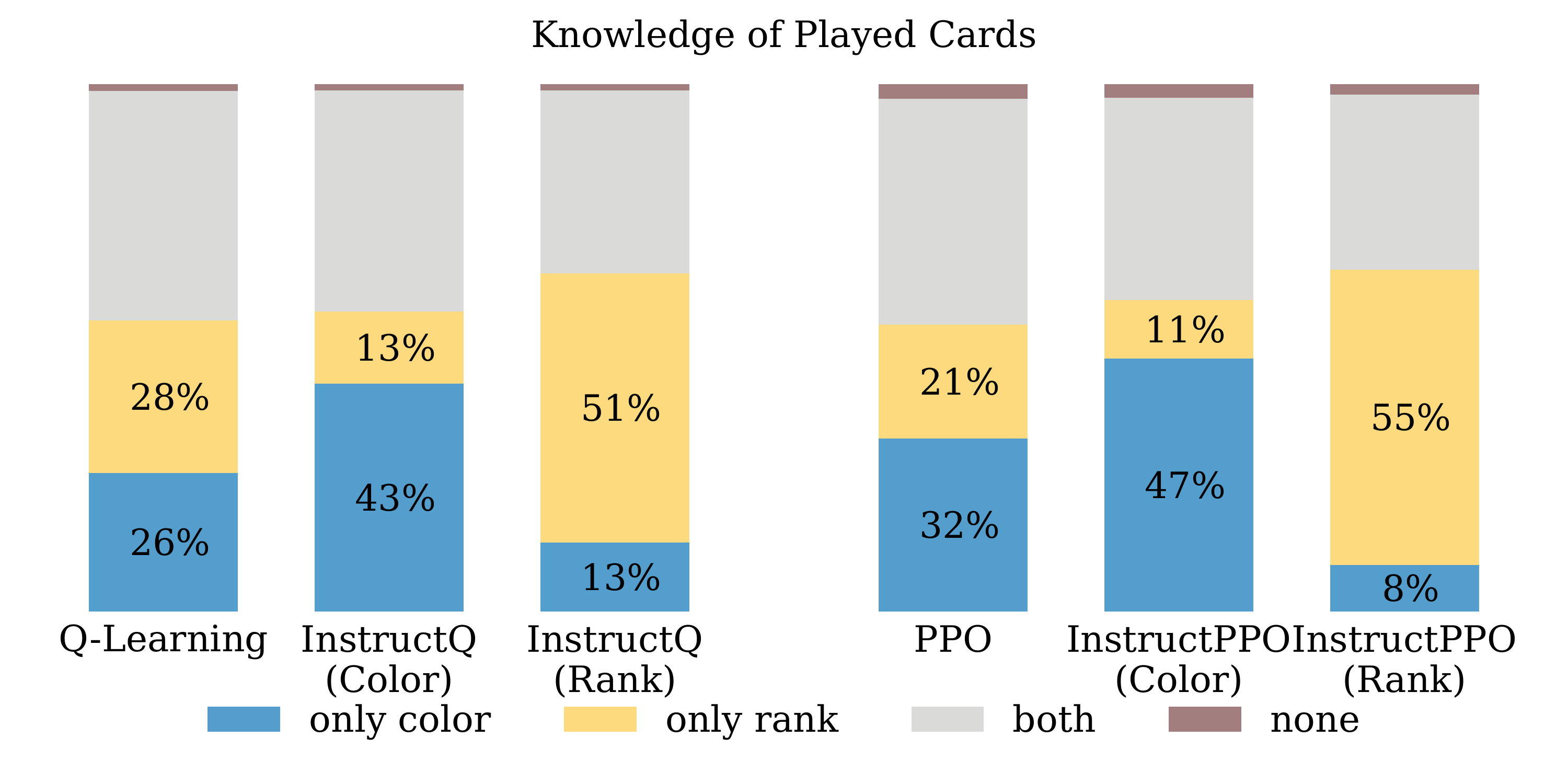}
    \vspace{-7mm}
    \caption{\small Knowledge of cards when the agent plays those cards. The knowledge of the cards is either revealed by hints or inferred public knowledge such as counting the remaining cards.  \textbf{Only color}: player knows the color but not the rank. \textbf{Only rank}: player knows the rank but not the color. \textbf{Both}: player knows exactly what the card is. \textbf{None}: player knows nothing about this card.}
    \vspace{-5mm}
    \label{fig:card_info}
\end{figure}

We then evaluate whether the policies are semantically different under different instructions and whether they satisfy the instructions. 
In Figure~\ref{fig:action_matrix}, we plot the conditional action matrix $p(a_{t+1} | a_{t})$ that shows the agent's response to partner's actions marginalized over time step $t$ in 1000 self-play games. All agents focus on hinting and playing the 5th (newest) card. The vanilla Q-learning agent uses a mixture of both color and rank hints to indicate playable cards. For instructQ and instructPPO, however, the agents heavily bias towards using the type of hints that they are instructed to use. 
This evidence alone is not sufficient to conclude that the agents' policies satisfy their instructions because the card played may have nothing to do with the hints.
In Figure~\ref{fig:card_info}, we check what kind of knowledge the agents have on the cards being played when they play a card. All agents primarily ($\geq 98\%$ of the time) play cards that they have knowledge about. The instruct agents rely significantly more on the type of hints that the instructions tell them to use. Videos of the games played by instructRL agents are available on the \href{https://sites.google.com/view/inst-rl-human-ai/home}{website} for more intuitive illustration of different policies.

\begin{figure}
    \centering
    \includegraphics[width=\linewidth]{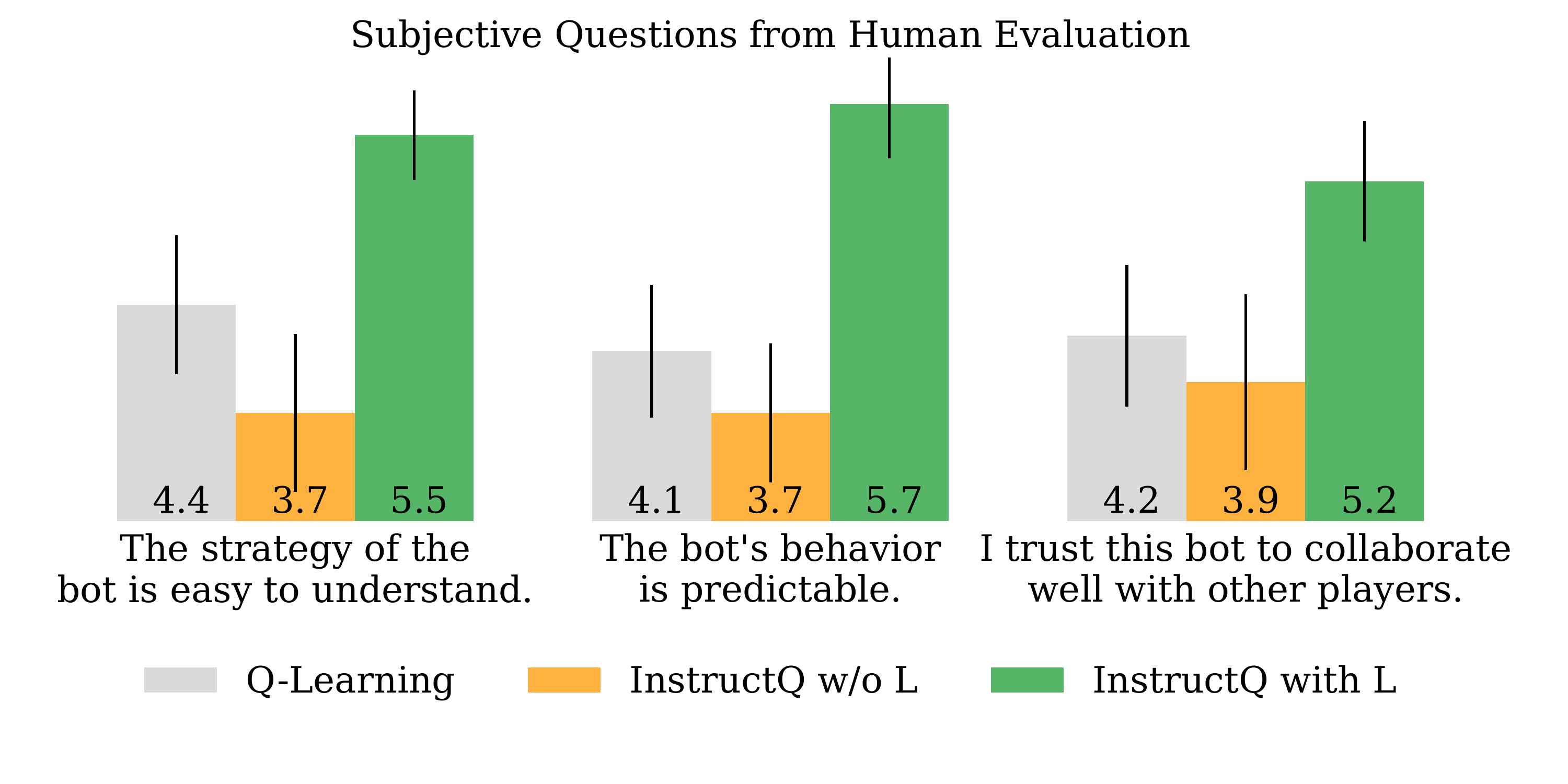}
    \vspace{-10mm}
    \caption{\small Human feedback on 3 subjective questions. The conclusion is consistent with that from the actual scores. Knowing the language instructions significantly improves the humans' experience when coordinate with the agents.}
    \vspace{-1mm}
    \label{fig:human-eval}
\end{figure}

\textbf{Human evaluation:} We ask 10 humans to evaluate the instructQ agents and the Q-learning baseline. Each participant plays with the Q-learning agent \emph{(Q-learning)}, one of the instructQ (color) or instructQ (rank) agent without seeing the corresponding language instructions \emph{(instructQ w/o L)}, and the remaining instructQ agent after seeing the instruction that it is trained to follow \emph{(instructQ with L)}. The results are shown in Table~\ref{tab:hanabi-human}. Without knowing the language, two instructQ agents get comparable results with the Q-learning baseline. However, when the language instructions are shown to the human partners, they achieve significantly higher scores and lose only 1 out of 10 games due to failed coordination. 
For the instructQ with language setting, we also ask the user two questions using 7-point Likert scale:
\begin{quote}
1) \emph{On the high level, the bot's strategy satisfies its instruction mentioned above.} \\
2) \emph{The language instruction helps me better collaborate with the bot.}
 \end{quote}
and the scores are 6.00 $\pm$ 0.24 and 6.20 $\pm$ 0.31 respectively. 
The results not only demonstrate that instructRL produces semantically different policies that follow the instructions, but also verify our observation that language communication is tremendously beneficial in human-AI coordination.
We also ask additional questions for all agents (shown in Figure~\ref{fig:human-eval}) and the results also confirm that knowing the instructions make the coordination experience much better.
Although the policies behind the orange and green bars are the same, humans think they are easier to understand, more predictable and trustworthy after seeing the language.

\begin{table}[]
    \centering
    \begin{tabular}{l c c c c}
    \toprule
    Method         & Score &  Game lost \\
    \midrule
    Q-learning      & 9.80 $\pm$ 3.35   & 5/10\\
    InstructQ w/o L & 7.80 $\pm$ 3.23   & 6/10 \\
    InstructQ with L   & \textbf{18.70 $\pm$ 2.18}  &  \textbf{1/10} \\
    \bottomrule
    \end{tabular}
    \caption{\small Human evaluation results. Each player plays with the Q-learning agent, one of the instructQ agent without knowing its language instruction and the other instructQ agent knowing its language instruction in \emph{random order}. \textbf{Score} shows mean $\pm$ standard error. \textbf{Game lost} shows how many games are terminated because all 3 lives are lost.}
    \label{tab:hanabi-human}
\end{table}

\textbf{Additional experiments on robustness and test-time adaptation:} In Appendix~\ref{app:robust}, we include additional experiments on the robustness of instructRL where we show that instructRL works well with less perfect prior policies generated with simpler, less prompt-engineered instructions. We also study how the performance deteriorates as we randomly corrupt the LLM prior.
In Appendix~\ref{app:adapt} we show that adding the LLM prior to the fixed Q-learning agents decreases the self-play performance of the agents while brings little improvement on their coordination performance with the corresponding instructQ agents, which encourages more future works along the line of \emph{test-time} adaption with language instructions.

\section{Conclusions}

In this paper we present instructRL, a framework for better human-AI coordination by training RL agents to follow natural language instructions that specify how the human would like to coordinate with AI. Instead of collecting labeled human data, we achieve this goal by using LLMs to construct prior policy conditioned on the instruction and use it to regularize RL to converge to the most desirable equilibria. 
Through both qualitative analysis and human evaluations, we show that instructRL converges to policies that satisfy the language instruction and humans can coordinate with RL agents much better if they are given the associated instructions used to train the agents. 

There are many exciting future directions, such as the problem of test-time adaption with language instructions detailed in Appendix~\ref{app:adapt}. As LLMs become more powerful, it will also be interesting to extend this work with more fine-grained instructions or enable users to specify instructions in more flexible ways, such as human-AI dialogues.

\section{Acknowledgments}

We would like to thank participants of the Hanabi human evaluation for playing with our agents We also thank the members of the Stanford ILIAD lab for their support and feedbacks during the development of this project and thank Samuel Sokota for discussions on different regularization techniques.
This works was supported by JP Morgan Faculty Award, DARPA YFA, AFOSR, ONR and NSF Award \#2006388 and \#2125511.

\newpage
\bibliography{reference}
\bibliographystyle{icml2023}

\newpage
\appendix
\onecolumn
\section{Implementation Details and Hyper-parameters}
\subsection{\toygame~Experiments}
\label{app:toy}
We use tabular Q-learning without neural networks for this experiment.
For hyper-parameters, we set $\epsilon=0.15$ for $\epsilon-${}Greedy exploration, $\lambda=0.25$ for instructQ regularization weight (only for Bob), $\texttt{lr}=0.02$ for learning rate. We use batch-size of 64 to reduce the variance of gradient estimation. For the vanilla Q-learning baseline, we simply set the $\lambda=0$ for Bob.

\subsection{Hanabi Experiments}
\label{app:impl-hanabi}

We implement instructQ and instructPPO based on the open sourced repository of off-belief learning (OBL)~\cite{hu2021off} and use their exact values for the existing hyper-parameters.
The implementation uses a parallel training setup with one training worker and thousands of rollout workers to speed up training.

Instead of training everything from scratch, we train all our models by fine-tuning an OBL level 1 policy.
OBL level 1 policy converges to an \textit{optimal grounded policy} where it only plays a card if it knows from the facts that the card is safe to play.
It uses a fair mixture of rank hints and color hints to tell the partner exactly which card to play.
The policy is a good start for fine-tuning because it does not contain any specific conventions.
We find three benefits for initializing with OBL level 1 policy. First, it significantly reduces the compute requirement for training because the agents do not need to relearn the entire game mechanics from scratch. Second, it makes multi-agent RL converge to the same policy across multiple runs with different seeds, improving the consistency and reproducibility of our method. Finally, it makes regularization easier since the agent uses both rank and color hints from the beginning.
We use Q-learning and PPO without any multi-agent heuristics to stay close to how OBL level 1 policy was originally trained. 

For the $\beta$ in $p_{\texttt{LLM}} = \textsc{Softmax}(\beta \cdot \text{logit})$, we set it to be 1 in instructQ and 2 in instructPPO experiments.
For instructQ, we set the regularization weight to $\lambda = 0.15$ initially and anneal $\lambda$ by half every 50K mini-batches. The policy is trained for a total of 250K mini-batches. 
Each mini-batch contains 128 episodes of games. 
For instructPPO, we set the regularization weight to $\lambda = 0.05$ initially and linearly anneal it by $0.008$ every 50K mini-batches until it reaches $0.01$ after 250K mini-batches. 
We then train for another 500K mini-batches with $\lambda=0.01$. For both Q-learning and PPO, we train their vanilla baselines using the same hyper-parameters but set $\lambda=0$.

\section{Illustration of Hanabi}
\label{app:hanabi-illu}
\begin{figure}[h]
    \centering
    \includegraphics[width=0.6\linewidth]{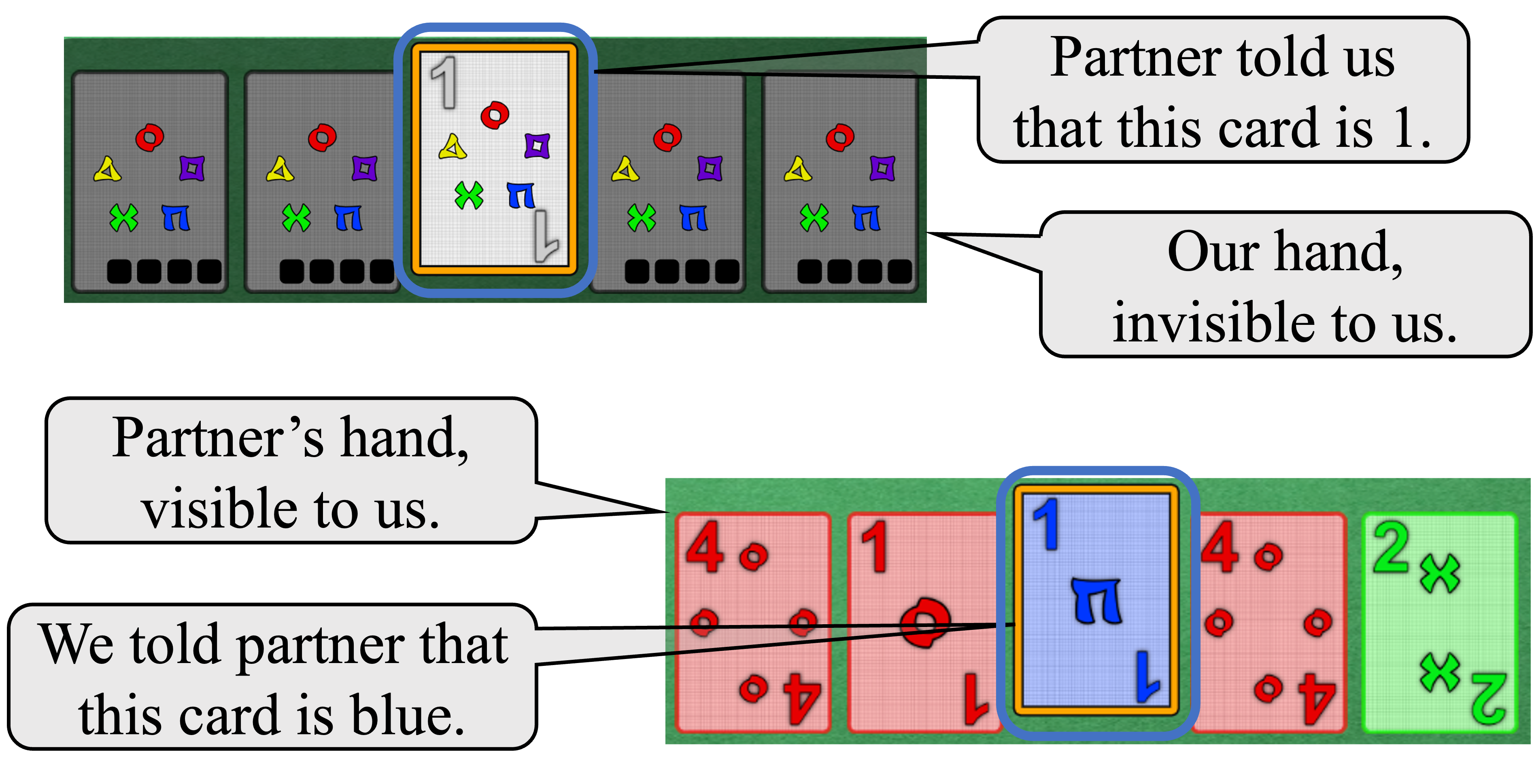}
    \caption{Illustration of Hanabi. In this example, our partner has previously told us that our 3rd card is 1 by hinting rank 1 to us. We also told our partner that they have a blue card by hinting blue to them. Note that if we had hinted rank 1 to our partner, both the red 1 and blue 1 would have been highlighted.}
    \label{fig:hanabi-illu}
\end{figure}

Figure~\ref{fig:hanabi-illu} illustrates an example time step in Hanabi game using the same interface that we use for human evaluation.
There are 5 different colors and 5 ranks in this game, although not all colors nor ranks appear in the current stage. First, we can see our partner's hand but not ours. We play cards based on the hints given to us by our partner.
In this example, our partner has previously told us that our 3rd card is 1 by hinting rank 1 to us. Meanwhile, we also know from the hint that all the other cards are \emph{not} 1s. We told our partner that they have a blue card by hinting blue to them. Note that if we had hinted rank 1 to our partner, both the red 1 and blue 1 would have been highlighted.
If this is the beginning of the game where no cards have been played, we should then play our 3rd card (the hinted one) because all the 1s are safe to play. Under a color-based policy, our partner will infer that the blue card we hinted at should also be a 1, and they will play it if the 1 we have is not blue and discard it otherwise.

\section{Robustness Analysis in Hanabi}
\label{app:robust}

In this section, we first measure the quality of the LLM priors in Hanabi by comparing them against a rule-based oracle.
Then we show that instructRL converges to nearly identical policies as the ones in the main paper when we swap in much simpler instructions that have not been prompt-engineered.
Finally, we examine how robust instructRL is in general by randomly corrupting the LLM priors.

\subsection{Accuracy of the LLM Priors}
\label{app:robust-acc}
First recall that our instruction used in the main text is \texttt{inst\_color} = \textit{``If my partner tells me the `color' of some of my cards, I should `play' those specific cards. If my partner does something else, e.g. discards their card or tells me the `rank' of my cards, then I may `hint color' to my partner"} to produce the color-based policy.
By swapping \textit{``rank"} for \textit{``color"} and \textit{``color"} for \textit{``rank"}, we get \texttt{inst\_rank} and use it produce the rank-based policy.
We can measure the ``accuracy" of the LLM prior by comparing the LLM's output with a rule-based oracle. For the \texttt{inst\_color}, the oracle returns 1 for all the actions that corresponds to 1) playing the hinted cards if partner's previous action is hinting color and 2) hinting color to the partner if partner's previous actions is not hinting color. It returns 0 for all other scenarios. The oracle for \texttt{inst\_rank} is similar. It turns out that the instructions mentioned above produce exactly the same results as their corresponding oracles, achieving $100\%$ accuracy. We refer to these instructions used in the main paper as \emph{original-inst}.

The original instructions use quotations in the instructions to emphasize the important keywords.
However, the quotations can be omitted with little cost on accuracy.
If we remove the quotations, the LLM makes 1 mistake out of the 3852 total questions for color and rank instructions respectively. The failure cases are shown in Figure~\ref{fig:failure-color} and Figure~\ref{fig:failure-rank} respectively.

\begin{figure}
\centering
\begin{minipage}{.5\textwidth}
  \centering
  \begin{quote}
\emph{Instruction: If my partner tells me the color of some of my cards, I should play those specific cards. If my partner does something else, e.g. discards their card or tells me the rank of my cards, then I should hint color to my partner.}

\emph{Previously: My partner told me that the color of my card at position E is blue.}

\emph{Question: Should I play my card at position D?}

\emph{Answer: \textcolor{red}{Yes (from LLM)}}

\textcolor{blue}{\emph{Expected Answer: No}}
  \end{quote}
  \caption{The failure case of \texttt{inst\_color} without quotations.}
  \label{fig:failure-color}
\end{minipage}%
\begin{minipage}{.5\textwidth}
  \centering
  \begin{quote}
\emph{Instruction: If my partner tells me the rank of some of my cards, I should play those specific cards. If my partner does something else, e.g. discards their card or tells me the color of my cards, then I should hint rank to my partner.}

\emph{Previously: My partner discarded their card at position E.}

\emph{Question: Should I hint color to my partner?}

\emph{Answer: \textcolor{red}{Yes (from LLM)}}

\textcolor{blue}{\emph{Expected Answer: No}}
  \end{quote}
  \caption{The failure case of \texttt{inst\_rank} without quotations.}
  \label{fig:failure-rank}
\end{minipage}
\end{figure}

\subsection{Performance with a Simpler Instruction}
\label{app:robust-simple}

We experiment with a simpler instruction template to test if our method is robust to errors made by LLMs.  
The color version of it is \texttt{inst\_color} = \emph{``If my partner told me the color of some of my cards, I should play those specific cards. 
Otherwise, I should hint color to my partner."} 
Similarly, we can obtain \texttt{inst\_rank} by swapping color for rank. We refer to these simpler instructions as \emph{simple-inst}.
The GPT-3.5 language model makes 113 mistakes under the simple color instruction and makes 110 mistakes under the simple rank instruction, roughly 3\% of the total query for both cases. 

Taking a closer look at the mistakes for the simple color instruction, we find that all of the mistakes happen when the previous action from the partner is not a color hint, i.e., when the previous action falls into the ``otherwise" category, and 109 out of the 113 mistakes appear when the previous action is a ``rank hint".
We show some of the mistakes in Table~\ref{tab:simple-inst-fail}. 
Although a human would not make similar mistakes even with this simple instruction, the mistakes from LLM are intuitive to explain because the simple instruction does not explicitly mention the possible types of actions when the partner does ``something else". Currently, this problem can be easily fixed by giving more details to the language model, as we do in the original instructions. As LLMs keep getting better, it may reach 100\% in the future even with the simple instruction.

\begin{table}[]
    \centering
    \begin{tabular}{p{7cm}  l   p{1cm}  p{1.5cm}}
    \toprule
    \centered{Previous action}  & Question  & LLM answer & Correct answer \\
    \midrule
        My partner played their card at position A. & Should I play my card at position B? &  Yes  & No \\
        My partner told me that the rank of my cards at position A and C is 2. & Should I play my card at position A? & Yes & No \\ 
        My partner told me that the rank of my cards at position A and D is 2. & Should I play my card at position A? & Yes & No \\
    \bottomrule
    \end{tabular}
    \caption{Examples of the failure cases when using a simple color-based instruction: \emph{``If my partner told me the color of some of my cards, I should play those specific cards. Otherwise, I should hint color to my partner."} The LLM answers are generated by GPT-3.5 model.}
    \label{tab:simple-inst-fail}
\end{table}

Next, we take the imperfect \emph{simple-inst} and use them in instructRL to see if the mistakes in the prior would cause the RL policies to diverge. 
We train instructQ with the \emph{simple-inst} using the exact same hyper-parameters as in our main paper. 
Then, we evaluate if the resulted color and rank policies are similar to their respective counterparts that use \emph{original-inst} by pairing them together in the cross-play setting. The results are show in Table~\ref{tab:perf-simple-inst}. 
We see that the cross-play scores between \emph{original-inst} and \emph{simple-inst} policies are nearly the same as the cross-play scores among \emph{original-inst}. 
Considering that Hanabi is a symmetric game and the player orders are randomized for each game, we can conclude that the instructQ policies trained with \emph{simple-inst} are nearly identical to the ones using more accurate priors obtained from complicated instructions. 
This experiment shows that our method is robust to imperfect priors and less prompt-engineered instructions, making it more generally applicable in the real world.

\begin{table}[]
    \centering
    \begin{tabular}{c c c c}
    \toprule
     Instruction Type & \emph{original-inst} $\times$~\emph{original-inst} & \emph{original-inst} $\times$~\emph{simple-inst} &  Performance drop \\
     \midrule
     Color & 23.80 $\pm$ 0.01  &  23.75 $\pm$ 0.03 & 0.05 (0.21\%) \\
     Rank  &  23.77 $\pm$ 0.01 & 23.71 $\pm$ 0.02 & 0.06 (0.25\%) \\
     \bottomrule
    \end{tabular}
    \caption{The cross-play scores of agents trained with the original instructions playing with agents trained with the simple instruction (\emph{original-inst} $\times$~\emph{simple-inst}) compared with the cross-play scores of the agents trained with the original instructions playing with agents trained with the same instruction but different seeds (\emph{original-inst} $\times$~\emph{simple-inst}). The small drop in the performance indicates that the agents trained with the imperfect \emph{simple-inst} is nearly identical as the ones trained with the perfect \emph{original-inst}.}
    \label{tab:perf-simple-inst}
\end{table}

\subsection{Performance with Randomly Corrupted Prior}
\label{app:robust-noise}

\begin{figure}
\centering
\hfill%
\begin{minipage}{.45\textwidth}
    \includegraphics[width=0.9\textwidth]{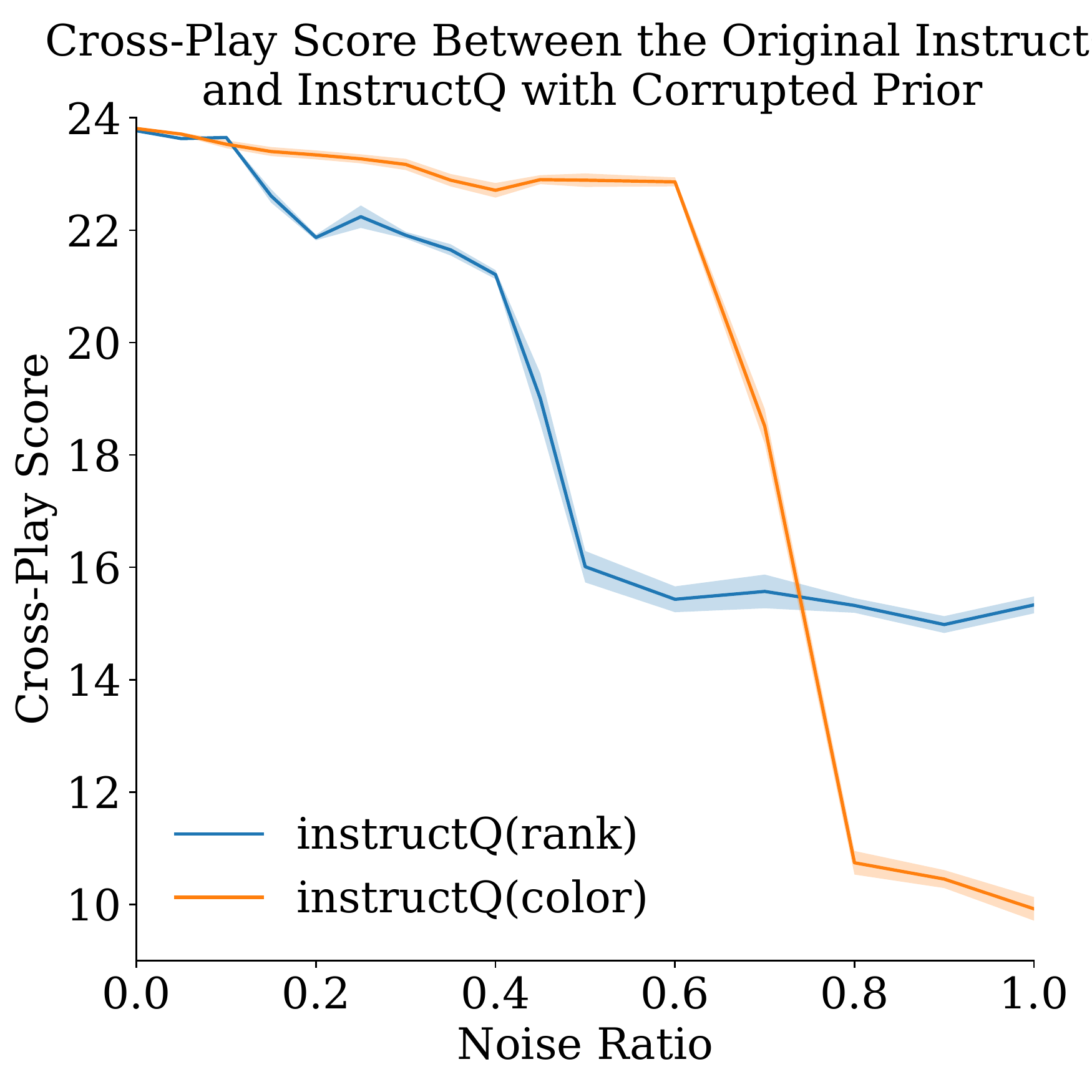}
    \caption{The cross-play scores between instructQ(noise=$x$) and instructQ(noise=0) for both color and rank LLM priors. The noise ratio $x$ controls the percentage of the prior values that are randomly flipped. The shaded area represents standard error.}
    \label{fig:noise}
\end{minipage}
\hfill%
\begin{minipage}{.45\textwidth}
    \includegraphics[width=0.9\textwidth]{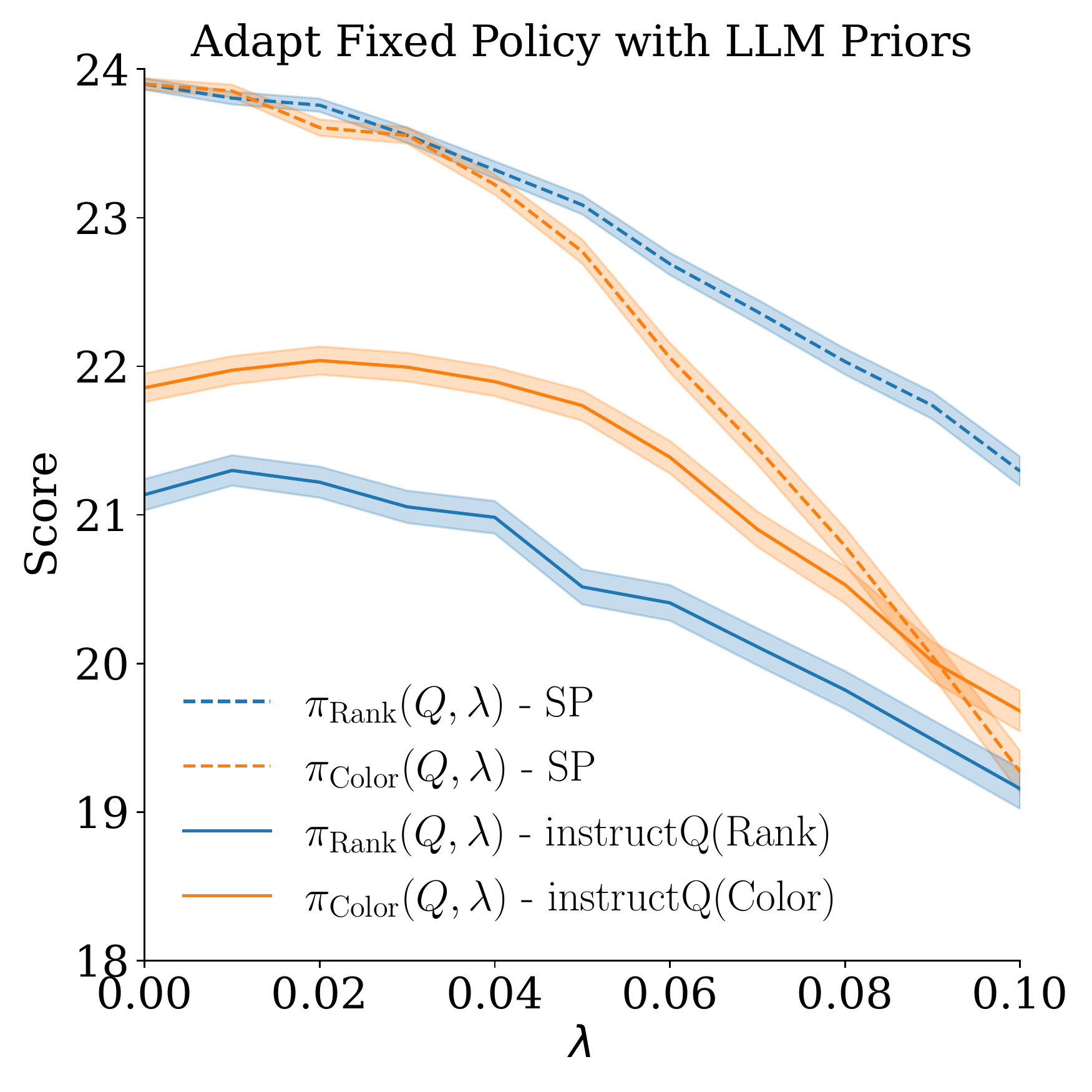}
    \caption{We augment a fixed Q-learning agent with color or rank LLM prior using different $\lambda$s. The figure shows their self-play performance (in dashed lines) and cross-play performance with their instructQ counterparts (in solid lines). The shaded area represents standard error.}
    \label{fig:adapt}
\end{minipage}
\hfill%
\end{figure}

Finally, we train instructRL with a spectrum of randomly corrupted prior policies. 
We start with 100\% accurate prior policies from \emph{original-inst} and randomly flip some values in the prior. 
We train instructQ(Color/Rank, $x$) where $x$ is the noise ratio ranging from 0 to 1. We evaluate how far the policies trained with corrupted prior diverge from instructQ(Color/Rank, $0$) by running cross-play between them.
The results are shown in Figure~\ref{fig:noise}. 
From the plot, we can see that,instructQ can withstand 10-15\% noise without noticeable drop in the cross-play scores for both color and rank instructions.
Then the performance starts to drop with different patterns where the cross-play scores of the rank policies drop faster than those of the color policy but they eventually stay at a higher level. Factors like the inductive bias of the learning algorithms, initialization and the annealing schedule of $\lambda$ may cause the curves to exhibit different patterns.

\section{Preliminary Experiments on Test-Time Adaptation in Hanabi}
\label{app:adapt}

An interesting future direction is to explore whether we can use language instructions to adapt a trained RL policy on-the-fly at test time. 
In this section we perform some preliminary studies to facilitate future exploration.

One way to formulate the problem is as follows. 
We want to coordinate with some unknown partner at test time and 
we do not have access to their exact policy but have access to the high level instruction that they used for training. 
The goal is to adapt our policy using the instruction and corresponding LLM prior so that it coordinates the partner better without any additional training.

We start with the Q-learning policy from our main paper as the base policy for ourselves, and use the two instructQ agents as unknown partners.
This is a simpler setting than the more general formulation defined above since all agents here are trained following the same setups except that our policy uses no KL-regularization.
We then modify our policy simply by adding the regularization term to the fixed Q-function without any additional fine-tuning, $\pi(Q, \lambda, p_\texttt{LLM}) = \argmax (Q + \lambda \log p_\texttt{LLM}$).
For convenience we denote the policies that use the color-based prior as $\pi_{\text{Color}}(Q, \lambda)$ and the ones that use the rank-based prior as $\pi_{\text{Rank}}(Q, \lambda)$.
We vary the values of $\lambda$ and plot both $\pi_{\text{Color}}(Q, \lambda)$ and $\pi_{\text{Rank}}(Q, \lambda)$'s self-play scores as well as the cross-play scores with their instructQ counterparts as a function of $\lambda$ in Figure~\ref{fig:adapt}. From the plot we see that the self-play performance drops monotonically for both $\pi_{\text{Color}}(Q, \lambda)$ and $\pi_{\text{Rank}}(Q, \lambda)$, indicating that adding the the priors post-hoc produces increasingly sub-optimal policies as $\lambda$ increases.
The cross-play scores, however, increase slightly for small $\lambda$s (from $21.13 \pm 0.11$ to $21.30 \pm 0.10$ for rank and from $21.86 \pm 0.10$ to $22.04 \pm 0.09$ for color). Although this indicates that they coordinate slightly better with the instructQ agents by adding the regularization to the fixed Q-function, both the quantity of improvement and the range of $\lambda$ that brings improvement is too small to be useful as it takes a large number of games to find the correct $\lambda$s.
This experiment emphasizes the importance of fine-tuning in our current algorithm since adding the LLM prior post-hoc leads to weaker policies without meaningful improvements for coordination.
Meanwhile, it also shows that test-time adaptation given instruction is hard and encourages more radical innovations.



\end{document}